\documentclass{article}
\usepackage{spconf,amsmath,graphicx,hyperref}

\usepackage{booktabs} 
\usepackage{multirow} 
\usepackage{array}  
\usepackage{adjustbox}
\usepackage{float}
\usepackage{placeins}
\usepackage{xspace}
\usepackage{makecell}
\usepackage{threeparttable}
\usepackage{xcolor}
\usepackage{graphicx} 
\usepackage{subcaption} 
\usepackage{caption} 
\usepackage{amssymb}

\title{What's on Your Plate? \\Inferring Chinese Cuisine Intake from Wearable IMUs}
\newcommand{\sysName}{CuisineSense\xspace}

\name{Jiaxi Yin,  Pengcheng Wang, Han Ding,  Fei Wang$^{*}$\thanks{$^*$Corresponding author.}}
  \address{Xi'an Jiaotong University, Xi'an, Shaanxi 710049, China 
     }

\begin{document}
%
\maketitle
\begin{abstract}
Accurate food intake detection is vital for dietary monitoring and chronic disease prevention. Traditional self-report methods are prone to recall bias, while camera-based approaches raise concerns about privacy. Furthermore, existing wearable-based methods primarily focus on a limited number of food types, such as hamburgers and pizza, failing to address the vast diversity of Chinese cuisine. To bridge this gap, we propose CuisineSense, a system that classifies Chinese food types by integrating hand motion cues from a smartwatch with head dynamics from smart glasses. To filter out irrelevant daily activities, we design a two-stage detection pipeline. The first stage identifies eating states by distinguishing characteristic temporal patterns from non-eating behaviors. The second stage then conducts fine-grained food type recognition based on the motions captured during food intake. To evaluate CuisineSense, we construct a dataset comprising 27.5 hours of IMU recordings across 11 food categories and 10 participants. Experiments demonstrate that CuisineSense achieves high accuracy in both eating state detection and food classification, offering a practical solution for unobtrusive, wearable-based dietary monitoring.The system code is publicly available at \url{https://github.com/joeeeeyin/CuisineSense.git}.

\end{abstract}


\section{Introduction}\label{sec:introduction}

Food intake detection, by accurately identifying users’ eating behaviors, provides a means to record dietary information. This enables timely adjustment of unhealthy eating habits and shifts health management from passive treatment toward proactive prevention. Such capabilities hold significant potential in health monitoring~\cite{kelly2018utilizing,tessier2025optimal}, weight management~\cite{min2017dietary}, and the prevention of prevalent chronic diseases such as hypertension and diabetes~\cite{marengoni2011aging,national1989diet}. Traditional self-report-based methods~\cite{bell2020automatic} for food intake detection are constrained by reliance on memory, subjective bias, and considerable time efforts, while camera-based approaches~\cite{bi2022eating} often raise privacy concerns. Recent studies have explored multimodal sensing \cite{ding2023mi}, e.g., IMU~\cite{thomaz2015practical, zhang2020necksense}, acoustic~\cite{kamachi2021automatic}\cite{ding2022ultraspeech}, piezoelectric~\cite{kalantarian2015monitoring}, for real-time, non-intrusive monitoring. While capable of classifying a closed set of foods, these approaches are highly susceptible to false positives triggered by common non-eating behaviors~\cite{vincent2010stacked,qing2023mar}.

To address these challenges, we present \sysName, a robust two-stage system leveraging IMUs from smart glasses and smartwatches. In the first stage, the intake state detection module (M1) models eating behaviors as the positive class in an anomaly detection framework, effectively distinguishing them from diverse non-eating states and filtering out irrelevant activities. In the second stage, the food type recognition module (M2) fuses complementary features: smartwatch signals capture fine-grained hand motion patterns such as utensil manipulation, while smart glasses capture head and jaw dynamics including postural shifts and chewing vibrations. This multimodal representation enables accurate recognition of food types in real-world scenarios. With its carefully designed architecture, \sysName achieves precise eating-action recognition using only a simple U-Net~\cite{ronneberger2015u} and a Swin Transformer~\cite{liu2021swin}, confirming the system’s feasibility.

To evaluate \sysName, we collected a data set of 11 Chinese foods (e.g. Baozi, dumplings, milk tea). It includes 27.5 hours of IMU recordings from 10 participants, with annotations for eating/non-eating states and food categories. Evaluation on this dataset shows \sysName achieves accurate food intake detection and fine-grained classification, enabling practical dietary monitoring.

The main contributions of this work are threefold:

(1) We design and develop \sysName, a wearable system that, for the first time, accurately recognizes up to 11 distinct categories of Chinese food, laying the groundwork for a practical, large-scale dietary monitoring solution.

(2) We propose a novel reconstruction-based anomaly detection framework for robust eating state detection. This approach effectively eliminates interference from diverse non-eating activities by identifying shifts in the data distribution of reconstructed sensor signals.

(3) We construct and release a comprehensive, richly annotated dataset of 27.5 hours of in-the-wild eating activities. Extensive experiments on this dataset validate the effectiveness and superior performance of \sysName.

\begin{figure*}[t]
    \centering
    \includegraphics[width=1\linewidth]{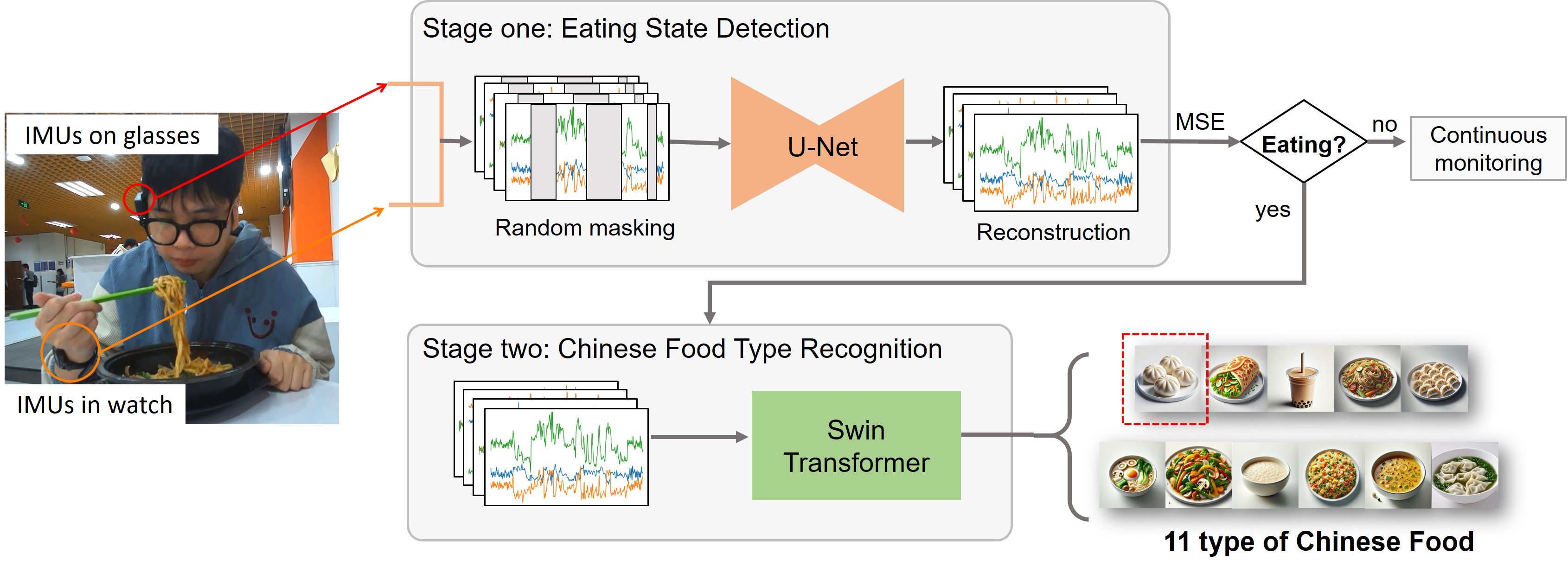}
    \caption{\sysName pipeline. A user wears a smartwatch and glasses, whose embedded IMUs capture hand and head motions, respectively. The IMU signals are first fed into the Eating State Detection module to determine if a food intake event is occurring. If yes, the signal segment is then passed to the second stage to classify the specific food being consumed.
}
    \label{fig:framework}
\end{figure*}

\section{\sysName}\label{sec:method}

\subsection{Rationale for Sensor Selection}\label{sec:sensor}
Inertial Measurement Units (IMUs), consisting of 3-axis accelerometers and gyroscopes, capture both linear acceleration $(a_x, a_y, a_z)$ and angular velocity $(\omega_x, \omega_y, \omega_z)$. These signals enable the inference of hand and head movements through changes in velocity, displacement, and orientation. In this study, we utilize the IMUs embedded in a smartwatch and smart glasses to record eating motion sequences.

\subsection{Framework Overview}\label{sec:framework-overview}

The framework of CuisineSense is illustrated in Fig. 1. A user typically wears a smartwatch and smart glasses, whose embedded IMUs capture hand and head motions. The IMU signals are first fed into the Eating State Detection module (M1) to determine if a food intake event is occurring. If an eating state is detected, the corresponding signal segment is then transferred to the Chinese Food Type Recognition module (M2), which classifies the specific food being consumed with higher reliability. Both modules employ standard yet effective backbones such as U-Net and Swin Transformer, showing that even without major network innovations, the lightweight two-stage hierarchical design yields excellent performance.

This two-stage design is crucial for filtering out diverse daily hand and head movements that could otherwise interfere with food classification. It prevents misclassifying non-eating activities into food categories, a common issue in single-stage systems. By incorporating this hierarchy, the approach ensures dietary logs remain accurate and trustworthy.

\subsection{Eating State Detection}\label{sec:eating-state-detection}

The first stage is designed to determine if a food intake event is occurring. A conventional approach would be to train a binary classifier with two classes of data: eating and non-eating. However, this is impractical due to the vast diversity of non-eating behaviors (e.g., writing, gesturing, screen-time). Collecting a comprehensive dataset that covers all possible non-eating scenarios is an exhaustive task, and any unseen non-eating activity could still lead to misclassification.

To address this challenge efficiently, we frame the eating state detection problem as anomaly detection. We adopt a classic reconstruction-based framework, which requires only the target ``normal" data for training—in our case, the eating activities. The model, trained through a mask-and-reconstruct pretext task, learns to effectively reconstruct the patterns of eating motions it was trained on. During inference, a given sequence is processed through this reconstruction network. If the reconstruction error is low, the sequence is classified as an eating event (normal). Conversely, a high reconstruction error indicates an anomaly, meaning the sequence is a non-eating activity that the model is not familiar with.

The core of the M1 module is a 1D U-Net, adapted from U-Net~\cite{ronneberger2015u}. The model consists of an encoder–decoder structure. The encoder extracts temporal features through stacked Conv1D–BN–ReLU blocks, where the number of channels doubles while the temporal resolution is halved after each convolution–pooling.

The decoder employs up-convolutions to restore temporal resolution and concatenates skip connections from the encoder to preserve low-level details. The final output reconstructs the input sequence, enabling reconstruction-based anomaly detection. To enhance feature learning, random masking is applied during training, forcing the model to infer missing segments and improve reconstruction fidelity. The reconstruction error is measured by mean squared error:
\begin{equation}\label{eq:mse}
 L_{recons} = \frac{1}{n} \sum_{i=1}^{n} (X_i - \hat{X}_i)^2
\end{equation}
Samples with errors above an empirically determined threshold are identified as non-eating, while those below are recognized as eating. This design enables robust detection even when non-eating behaviors are diverse and not exhaustively observed during training.

\subsection{Chinese Food Type Recognition}\label{sec:food-type-recognition}

Since different food types involve distinct patterns and speeds of hand and head movements, this stage must capture both short- and long-term temporal features from the eating sequence. To this end, we employ a 1D Swin Transformer~\cite{liu2021swin}, leveraging its hierarchical structure and shifted window mechanism to learn diverse temporal dependencies and achieve efficient, fine-grained food classification.

Specifically, the input IMU food intake sequence is first partitioned into patches and linearly embedded before being processed by stacked Swin Transformer blocks. Unlike standard multi-head self-attention (MSA), the Swin Transformer adopts a shifted window self-attention mechanism (W-MSA and SW-MSA) to efficiently capture both local and long-range temporal dependencies. The model outputs an 11-class probability vector corresponding to the food types. The network is trained using the cross-entropy loss function:
\begin{equation}
\mathcal{L}(\hat{y}, y) = -\sum^{C}_{i=1}y_i \log(\hat{y}_i)
\end{equation}
where $C$ is the number of food categories, $y_i$ is the one-hot encoded ground truth label, and $\hat{y}_i$ is the predicted probability. This formulation enables the model to learn discriminative temporal-motion features from multi-sensor IMU data and accurately recognize food categories.

\section{experiments}\label{sec:experiments}

\subsection{Data Collection}\label{data-collection}

We collected a dedicated dataset with 10 volunteers (6 males, 4 females, BMI 16.3–27.5). Each participant wore a Samsung Gear Sport smartwatch on their dominant hand (50 Hz) and a pair of glasses with an IMU (10 Hz) to capture hand motions and chewing-related head vibrations. Data were transmitted via Wi-Fi to a laptop for storage.

The dataset covers three utensils (chopsticks, spoon, hand) and 11 Chinese food categories (Mixed Noodles, Dumplings, Noodle Soup, Stir-fry, Baozi, Pancake, Milk Tea, Congee, Fried Rice, Soup, Wontons) in natural scenarios. Each volunteer contributed ~15 minutes per category, totaling 27.5 hours (15 min×11 foods×10 subjects). For accurate annotation, videos recorded the process, with timestamps syncing data streams. Videos were used to manually segment eating versus non-eating actions.

Based on our observation that a complete eating gesture averaged 10.17 seconds, we segmented the continuous data streams into 10.24-s windows to encapsulate full gesture cycles. A sliding window of 2.56s with 50\% overlap was applied to augment sequences. For model training, we adopted a subject-independent split: from each volunteer’s 2700 frames, 300 middle frames were reserved for testing and the rest for training, yielding 70,912 training and 10,496 test samples. After segmentation, we obtained 16,625 eating and 22,013 non-eating sequences. Each sample includes synchronized smartwatch (128×6) and glasses (25×6) IMU data, paired with labels for food intake state and food category. This dataset provides a solid foundation for training and evaluation.

\subsection{Implementation}\label{sec:implementation}
All experiments were conducted on a high-performance server equipped with four NVIDIA GeForce RTX 3090 GPUs (CUDA 11.7). Models were implemented in Python 3.8 using PyTorch 2.0.1, and trained with the Adam optimizer\cite{kingma2014adam} (batch size 16, learning rate 1e-4) for 100 epochs with shuffled samples at each epoch.

\subsection{Results}\label{sec:results}

\begin{figure}[t]
    \centering
    \includegraphics[width=0.85\linewidth]{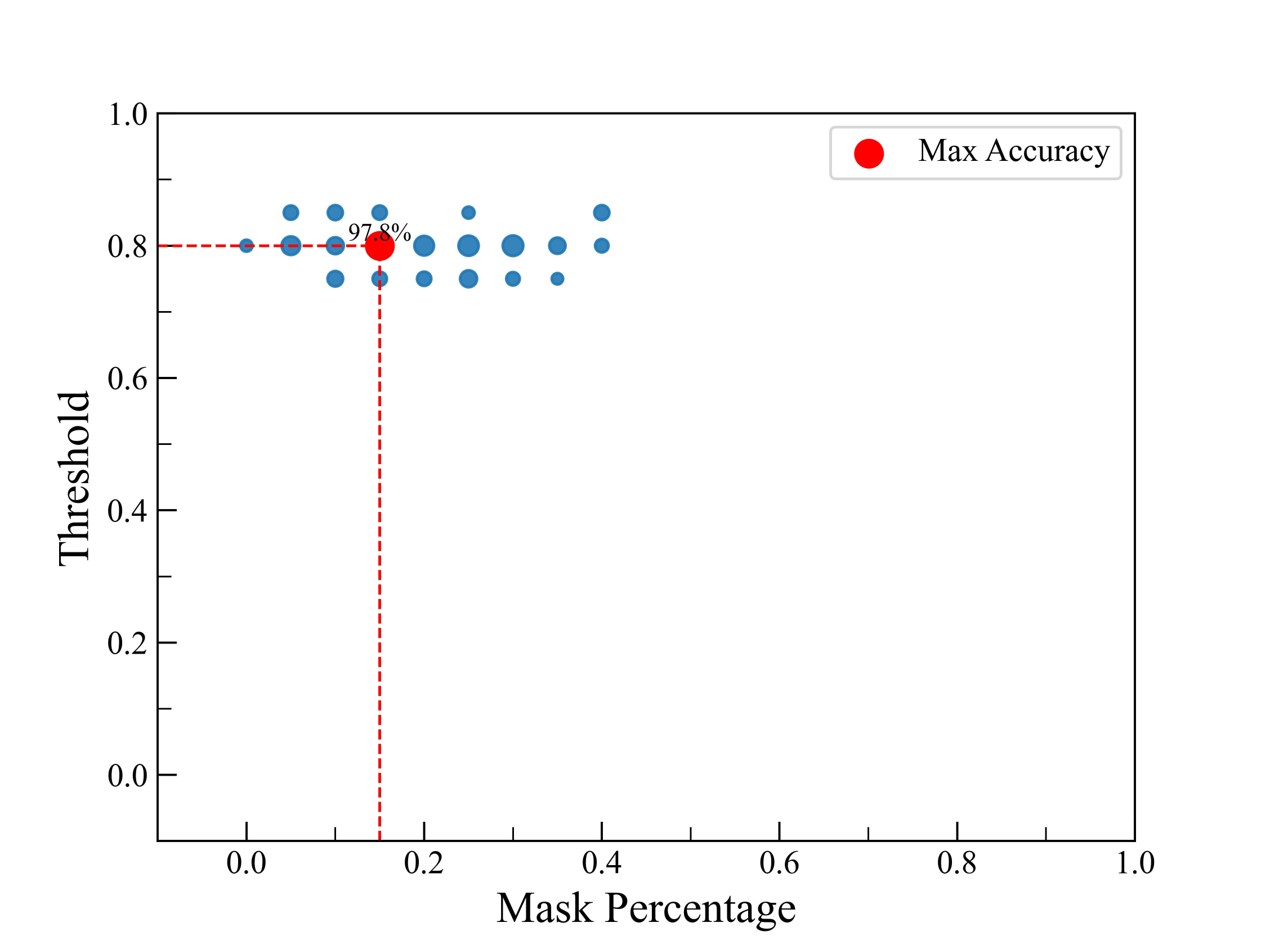}
    \caption{Hyperparameter search for the eating state detection module. The plot shows the top 20 masking ratio–MSE threshold combinations, with circle size proportional to accuracy. Best performance was achieved with a mask ratio of 0.15 and an MSE threshold at the 80th percentile of the losses.}
    \label{fig:m1-mask-threshold}
\end{figure}

\textbf{(1) Eating State Detection Performance.} The first stage of \sysName distinguishes eating actions from non-eating movements that occur within a meal period. The classification is based on the reconstruction error of the IMU signals after a masking operation~\cite{haresamudram2020masked}. We conducted extensive experiments to tune the hyperparameters, exploring various masking ratios and MSE thresholds for reconstruction error. Figure~\ref{fig:m1-mask-threshold} illustrates the top 20 hyperparameter combinations, where a larger circle indicates a better accuracy.

The best performance was achieved when the masking ratio was set to 0.15 and the lowest 80\% in MSE were selected. With these parameters, our 1D U-Net model~\cite{ronneberger2015u} reached an accuracy of 97.94\% in separating eating from non-eating states. This significantly outperforms a standard autoencoder~\cite{hinton2006reducing}, which only achieved 83.39\%. Therefore, we selected U-Net as the backbone for our first-stage module.

\textbf{(2) Food Type Recognition Accuracy.} Once the U-Net in the first stage identifies an eating action, the second stage classifies its food type. For this task, we evaluated two representative backbone architectures: a 1D ResNet~\cite{he2016deep} and a 1D Swin Transformer~\cite{liu2021swin}. As shown in Table~\ref{tab:food-type-recognition-backbone}, the network combining U-Net with the Swin Transformer achieved superior performance, reaching a classification accuracy of 88.44\% on Chinese food intake. This result demonstrates that the Swin Transformer can more effectively learn the complex short- and long-term temporal features inherent in the process of eating Chinese food. Furthermore, thanks to the lightweight nature of the 1D network, the model is highly efficient, with an inference time of just 2.16$\pm$0.02 ms for each 10.24-second window, demonstrating comparable efficiency on both GPU (NVIDIA 3090) and CPU, demonstrating its potential for real-time applications.


\begin{table}[t]
\centering
\caption{Food type recognition accuracy.}
\begin{tabular}{l l}
\hline
Network Combination & Accuracy\\
\hline
U-Net+ResNet~\cite{he2016deep} & 80.31\% \\
U-Net+Swin Transformer~\cite{liu2021swin} & 88.44\% \\
\hline
\end{tabular}
\label{tab:food-type-recognition-backbone} 
\end{table}

\begin{figure}[t]
    \centering
    \includegraphics[width=0.9\linewidth]{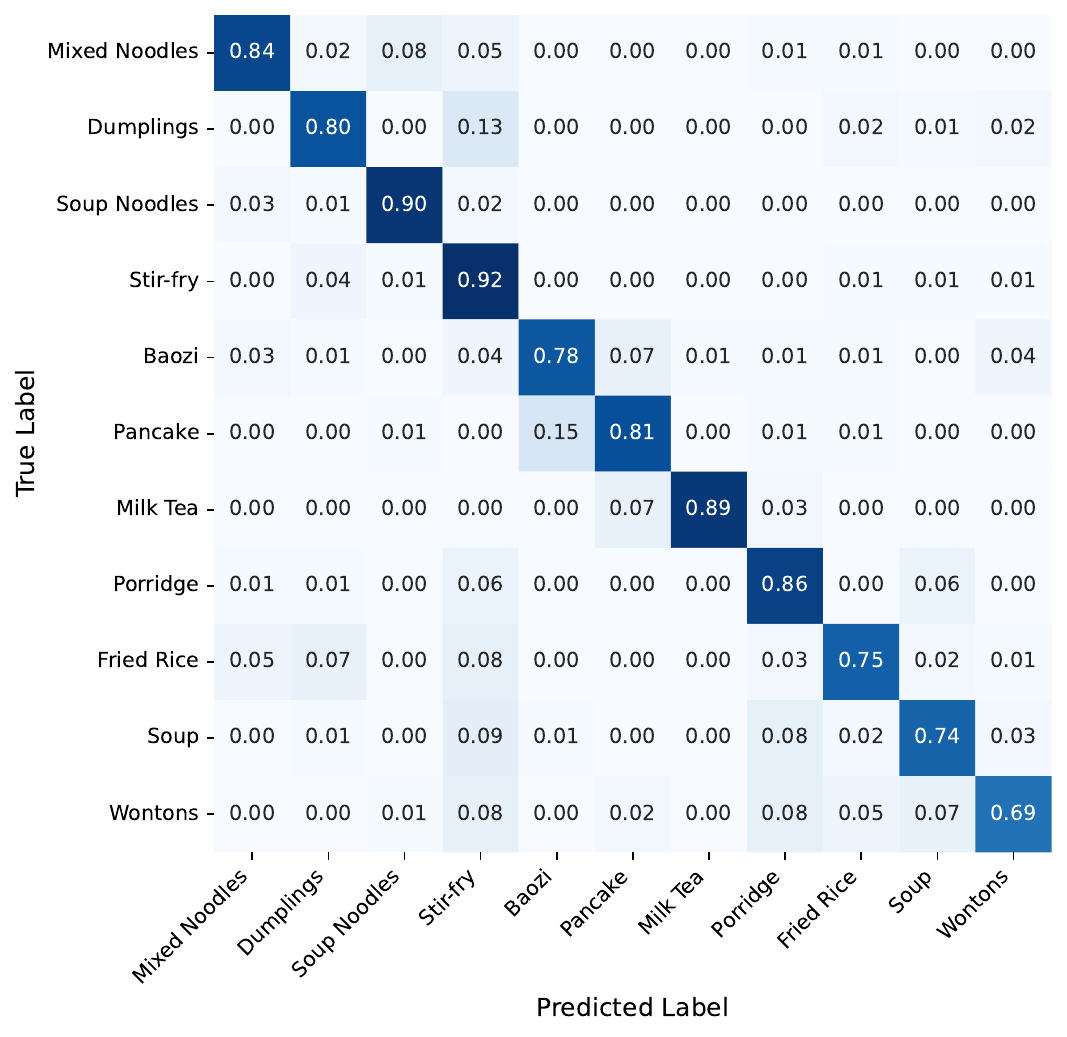}
    \caption{Confusion matrix of food type recognition.}
    \label{fig:confusion-matrix}
\end{figure}

\textbf{(3) Confusion Matrix of Food Type Recognition.} The confusion matrix in Fig.~\ref{fig:confusion-matrix} details the correct and incorrect classifications for the food type recognition stage. From the matrix, we can observe several key findings. Stir-fried dishes achieve the highest accuracy (92\%) due to their distinct chopstick motions but are sometimes confused with dumplings, likely due to similar chewing patterns. Baozi and pancakes show some confusion, which can be attributed to similar hand-to-mouth gestures. Wontons have the lowest accuracy (69\%) and are most often misclassified as congee, fried rice, or soup, likely because they are all eaten with a spoon and may share less distinct temporal features.

\textbf{(4) Ablation Study on the Effectiveness of the First Stage.} To validate the effectiveness of our two-stage design, we conducted an ablation study by removing the first-stage U-Net module. In this alternative setup, the second-stage classifier's output score is used directly to determine the eating state. Specifically, if the confidence score for all food categories falls below a certain threshold, the action is classified as non-eating. Otherwise, it is classified as the food type with the highest score.


As shown in Table~\ref{tab:threshold_horizontal}, we tested thresholds from 0.1 to 0.9. For instance, with a threshold of 0.5, the accuracy is merely 45.54\%. Even with the optimal threshold of 0.9, the peak accuracy only reaches 52.43\%. This is substantially lower than the 88.44\% accuracy achieved by our complete two-stage system (shown in Table~\ref{tab:food-type-recognition-backbone}). This significant performance drop demonstrates that the first stage is crucial for filtering out non-eating activities and confirms the effectiveness of our proposed two-stage methodology.

\begin{table}[t]
\small
\caption{Food type recognition without the first stage under different classification score threshold~(CST).}
\centering
\setlength{\tabcolsep}{1pt} 
\begin{tabular}{lccccccccc}
\hline
ST & 0.1 & 0.2 & 0.3 & 0.4 & 0.5 & 0.6 & 0.7 & 0.8 & 0.9 \\
\hline
Acc.(\%) & 45.88 & 45.88 & 45.88 & 46.07 & 45.54 & 47.76 & 49.29 & 50.61 & 52.43 \\
\hline
\end{tabular}
\label{tab:threshold_horizontal} 
\end{table}

\section{Conclusion}\label{sec:conclusion}

We introduced \sysName, a novel two-stage system for recognizing Chinese food intake using IMU sensors embedded in a smartwatch and smart glasses. The proposed system first employs a U-Net-based anomaly detection module to robustly identify eating states, followed by a 1D Swin Transformer for fine-grained classification of 11 different Chinese food categories. Our extensive experiments, conducted on a newly collected 27.5-hour dataset, validate the effectiveness of our approach.
In summary, \sysName represents a significant step toward an accurate, unobtrusive, and privacy-preserving solution for automated dietary monitoring, with promising implications for proactive health management.




{
\small
\bibliographystyle{IEEEbib}
\bibliography{strings,refs}
}
\end{document}